\documentclass{article} 
\usepackage{iclr2025_conference,times}

\usepackage{amsmath,amsfonts,bm}

\def\eqref#1{equation~\ref{#1}}

\def\1{\bm{1}}

\DeclareMathAlphabet{\mathsfit}{\encodingdefault}{\sfdefault}{m}{sl}
\SetMathAlphabet{\mathsfit}{bold}{\encodingdefault}{\sfdefault}{bx}{n}

\usepackage{multirow}
\usepackage{hyperref}
\usepackage{url}
\usepackage{tabularx}
\usepackage{threeparttable}
\usepackage{booktabs}
\usepackage{graphicx}
\usepackage{enumitem}
\usepackage{siunitx}

\title{VISTAR:A User-Centric and Role-Driven Benchmark for Text-to-Image Evaluation}

\author{%
	Kaiyuan Jiang\thanks{Equal contribution.}\thanks{Work done during an internship at LinkSure.} \\ 
	Peking University \\
	\texttt{kyjiang@stu.pku.edu.cn}
	\And
	Ruoxi Sun\footnotemark[1] \\ 
	LinkSure \\
	\texttt{sunrx@zenmen.com}
	\And
	Ying Cao \\
	LinkSure \\
	\texttt{caoying01@zenmen.com}
	\AND 
	Yuqi Xu\footnotemark[2] \\ 
	University of Glasgow \\
	\texttt{1640732771@qq.com}
	\And
	Xinran Zhang\footnotemark[2] \\
	Boston University \\
	\texttt{xinranz1208@gmail.com}
	\And
	Junyan Guo \\
	LinkSure \\
	\texttt{guojy01@zenmen.com}
	\AND 
	ChengSheng Deng \\
	LinkSure \\
	\texttt{dengcs@zenmen.com}
}

\iclrfinalcopy 

\begin{document}

\maketitle

\begin{abstract}
We present \textbf{VISTAR}, a user-centric, multi-dimensional benchmark for text-to-image (T2I) evaluation that addresses the limitations of existing metrics—namely narrow dimensional coverage, lack of task specificity, and the tension between reproducible physical quantification and abstract semantic understanding.  
VISTAR introduces a \emph{two-tier hybrid paradigm}: deterministic, scriptable metrics for physically quantifiable attributes (text rendering, lighting consistency, geometric coherence) and a novel \emph{Hierarchical Weighted P/N Questioning} (HWPQ) scheme that constrains large vision–language models within a structured, adversarial-question framework to assess abstract semantics (style fusion, cultural fidelity, material accuracy).
Grounded in a Delphi study with 120 domain experts, we define seven representative user roles and nine evaluation angles, identifying 28 critical role–aspect pairs via Bayesian optimisation.  
The benchmark comprises 2,345 structured prompts (\texttt{VISTAR-Core}) and 500 adversarial prompts (\texttt{VISTAR-Hard}), validated on 15,000+ human pairwise comparisons.  
All metrics exceed a 75 \% alignment threshold; HWPQ achieves 85.9 \% mean accuracy on abstract angles, outperforming conventional VQA baselines by over 7 \%.
Comprehensive evaluation of state-of-the-art models reveals no universal champion; role-weighted scores reorder rankings and provide actionable guidance for domain-specific deployment.  
Ablation studies confirm the necessity of every design component.  
Code, data, and full questionnaires are released to foster reproducible T2I assessment and model improvement.
\end{abstract} 

\section{Introduction}

Recent breakthroughs in text-to-image generation have demonstrated a remarkable capability to produce diverse, high-fidelity images from natural language prompts. Particularly, the synergistic integration of Vision-Language Models (VLMs) and Large Language Models (LLMs) has significantly enhanced image generation model's ability to comprehend complex semantics, parse instructional structures, and execute fine-grained control. This technological evolution is powerfully driving the widespread application of generative models across various domains, including creative design, assisted drawing, and virtual reality.

Simultaneously, to scientifically evaluate the performance of image generation models, researchers have proposed diverse assessment methodologies. These include the Fréchet Inception Distance, which measures the distance between generated and real image distributions \citep{heusel2017gans}; text-to-Image Alignment metrics, such as CLIPScore \citep{hessel2021clipscore}, quantifying semantic consistency between prompt and image, and fine-grained evaluation mechanisms focusing on detail fidelity and attribute binding. While these methods offer foundational performance feedback for model development, they generally suffer from limited evaluation dimensions, insufficient discriminative power, and a lack of task specificity. This makes it challenging to comprehensively reflect a model's performance across varied real-world application scenarios.

Notably, the evaluation frameworks developed for LLMs offer significant inspiration for assessing image generation models \citep{wang2024mmlu}. Language model evaluations typically span multiple task dimensions, such as mathematical computation, code generation, scientific reasoning, and language writing. This systematic approach effectively reveals a model's specific performance across different capability levels. Such evaluations not only enhance the interpretability of model behavior but also provide clear criteria for users selecting models for diverse application contexts.

Inspired by this, we propose VISTAR (Visual STAndards for T2I Roles), a multi-dimensional evaluation benchmark for text-to-image generation tasks constructed with a user-role-centric perspective. VISTAR is the first framework to systematically address the existing gap in multi-dimensional evaluation of text-to-image generation capabilities. It aims to overcome the limitations of current evaluation systems regarding capability coverage, task relevance, and result operability. By thoroughly considering the specific concerns of different user groups, VISTAR establishes a more discriminative and practically guiding evaluation system. It comprehensively aligns with actual application needs and user task orientations, precisely characterizing model performance in various role-specific contexts. VISTAR not only provides structured, fine-grained evaluation criteria for generative models but also offers researchers effective guidance for identifying research priorities and pinpointing capability bottlenecks, thereby promoting the deployment and optimization of image generation technology in more diverse scenarios.

Inspired by this, we introduce \textbf{VISTAR} (Visual STAndards for T2I Roles), a multi-dimensional, user-centric evaluation benchmark for text-to-image generation. VISTAR addresses the critical gaps in existing evaluation methodologies by shifting the focus from generic quality scores to role-specific, actionable insights. Our framework is built upon a systematically engineered ontology of seven user roles (e.g., Graphic Designer, Storyboard Artist) and nine evaluation angles, derived from an expert-led Delphi study. This design ensures that our evaluation is not only comprehensive but also directly relevant to real-world application scenarios.
The main contributions of this work are three-fold:
\begin{enumerate}[leftmargin=*,topsep=3pt,itemsep=0pt]
	\item \textbf{A Novel Role-Centric Evaluation Paradigm.} We are the first to propose a ``user role-evaluation aspect'' dual-dimensional framework. This paradigm enhances the interpretability and task-relevance of T2I evaluation, providing a structured basis for users to select models that best fit their specific needs.
	
	\item \textbf{A Hybrid and Fine-Grained Metric System.} We construct a two-tier evaluation mechanism. For physically quantifiable attributes like text rendering and spatial consistency, we design deterministic, scriptable metrics. For abstract semantics such as style fusion and cultural fidelity, we introduce the Hierarchical Weighted P/N Questioning (HWPQ) scheme, a novel method that constrains a VLM's reasoning for robust and reproducible assessment.
	
	\item \textbf{A Comprehensive Benchmark and SOTA Analysis.} We release \texttt{VISTAR-Core}, a benchmark of 2,345 structured prompts, and an accompanying human preference dataset. Our extensive evaluation of state-of-the-art models reveals nuanced capability trade-offs, demonstrating that model rankings shift based on user roles and providing the community with a valuable decision-support tool.
\end{enumerate}

We believe that the multi-dimensional, high-fidelity evaluation framework proposed by VISTAR will offer a new research perspective for the text-to-image generation field, enabling researchers to optimize model performance more targetedly and driving the domain towards more practical and controllable development.

\section{Related Works}
\label{gen_inst}

Since 2016, image generation models have attracted sustained attention from the research community. Early approaches relied on deep neural networks, and the subsequent advent of generative adversarial networks \citep{goodfellow2014generative} accelerated their proliferation. The decisive breakthrough in text-to-image (T2I) generation arrived with autoregressive models such as DALL·E \citep{ramesh2021zero} and diffusion-based frameworks exemplified by Stable Diffusion\citep{rombach2022high}, inaugurating an era characterized by high-fidelity outputs and accurate semantic alignment. Subsequent iterations of diffusion architectures—particularly those incorporating flow matching\citep{lipman2022flow}—have substantially elevated sampling efficiency. Concurrently, enhanced text encoders and refined cross-modal feature alignment have continued to drive perceptual quality upward. Recent exemplars, including the GPT series\citep{heusel2017gans}, SeeDream\citep{gao2025seedream}, and FLUX, now produce imagery that is visually indistinguishable from photographs.

Nevertheless, in practical applications such as portrait customization, artistic design, or poster production, existing models still confront multifaceted challenges: fine-grained controllability, domain-specific semantic understanding, and output stability. Consequently, a systematic and precise delineation of the capability boundaries of current models is essential for directing future optimization and exploratory research.

Early evaluation protocols leveraged off-the-shelf datasets (e.g., COCO \citep{lin2014microsoft}, CUB-200-2011 \citep{wah2011caltech}) and their associated captions, focusing on generation quality under limited-sample regimes. As T2I models matured, evaluation paradigms shifted toward more demanding dimensions. DrawBench \citep{saharia2022photorealistic} and RichHF \citep{liang2024rich} introduced linguistically complex prompts to probe text–image alignment.

To enable finer-grained capability profiling, HE-T2I \citep{petsiuk2022human} introduced sub-benchmarks spanning numerosity, shape accuracy, and facial fidelity; DALL-EVAL \citep{cho2023dall}assessed visual reasoning, caption consistency, image quality, and social bias. GenEval \citep{ghosh2023geneval} emphasized object-detection-oriented metrics; DSG \citep{cho2023davidsonian} and DPG-bench \citep{hu2024ella} analyzed semantic alignment via scene graphs; T2I-CompBench++ \citep{huang2025t2i} stressed robust attribute binding. Arena-style platforms such as artificialanalysis.ai further incorporated human-preference rankings to enhance subjective interpretability.

While these benchmarks have advanced the field, they share a common limitation: a \textbf{model-centric} perspective that prioritizes instruction following over \textbf{user-centric} utility. This leaves a critical gap in understanding real-world applicability. We address this gap directly by presenting the first role-centric evaluation suite.By explicitly foregrounding user-role alignment and real-world utility, our benchmark extends prior capability dimensions whilemaintaining rigorous granularity (see Table~\ref{tab:benchmark_comparison}).

\begin{table}[htbp]
	\centering
	\setlength{\tabcolsep}{1pt} 
	\begin{threeparttable}
		\caption{Comparison of Text-to-Image Evaluation Benchmarks} 
		\label{tab:benchmark_comparison}
		
		\begin{tabular}{@{}l c l c l l l@{}} 
			\toprule
			
			\textbf{Benchmark} & \textbf{Role-Centric} & \textbf{Eval. Paradigm} & \textbf{Dims} & \textbf{Metric Granularity} & \textbf{Primary Output} & \textbf{Open Source} \\
			\midrule 
			DrawBench     & $\times$ & HP\tnote{a}   & N/A & Overall Score   & Relative Ranking & Yes (Prompts) \\
			T2I-CompBench & $\times$ & DM\tnote{b}   & 5   & Attribute-level & Numerical Scores & Yes \\
			DALL-EVAL     & $\times$ & D+H\tnote{c}  & 4   & Aspect-level    & Numerical Scores & Yes \\
			T2I-Arena     & $\times$ & HP\tnote{a}   & N/A & Overall Score   & Elo Ranking      & Partial \\
			\midrule 
			\textbf{VISTAR (Ours)} & $\checkmark$ & H(DS)\tnote{d} & 9   & Role \& Angle-level & Numerical Scores & Yes \\
			\bottomrule 
		\end{tabular}
		\begin{tablenotes}[para,flushleft]
			\small 
			\item[a] HP: Human Preference 
			\item[b] DM: Deterministic Metrics 
			\item[c] D+H: Deterministic + Human 
			\item[d] H(DS): Hybrid (Deterministic + Semantic) 
		\end{tablenotes}
	\end{threeparttable}
\end{table}

Current T2I evaluation metrics fall into three categories: (i) image realism scores, (ii) text–image alignment scores, and (iii) multimodal large language model (MLLM)–based scores. Early work adopted Fréchet Inception Distance , Inception Score \citep{salimans2016improved} , and LPIPS \citep{zhang2018unreasonable}to quantify perceptual fidelity to real images; however, these metrics inadequately capture instruction-following capacity.

To address this limitation, CLIP-based \citep{radford2021learning} and refined BLIP-based \citep{li2022blip} similarity scores were introduced to measure semantic alignment between prompts and generated images. Subsequent studies fine-tuned BLIP-2 \citep{li2023blip} variants on human preference data to assess aesthetic quality more directly. Recent work further employs large language models (LLMs) to generate multi-turn questions, followed by vision-language models (VLMs) performing visual question answering (VQA) \citep{liu2023g}to yield interpretable, fine-grained diagnostics.Nevertheless, these scoring mechanisms, while valuable, remain largely holistic and generalist. They typically yield a single score for abstract concepts like "alignment" or "quality", lacking the \textbf{diagnostic granularity} needed to pinpoint specific model failures. Furthermore, their application is often uniform across all use cases, failing to account for the \textbf{task-specific} needs of different users. To overcome this limitation, we craft tailored metrics for each evaluation angle and validate their perceptual alignment with human judgments, thereby substantiating the reliability and interpretability of the proposed framework.

\section{Method}
\label{headings}

\subsection{Core Principles: Roles, Aspects, and Quality Alignment}
\label{ssec:core_principles}
\begin{figure}[htbp]
	
	\centering

	\includegraphics[width=0.6\linewidth]{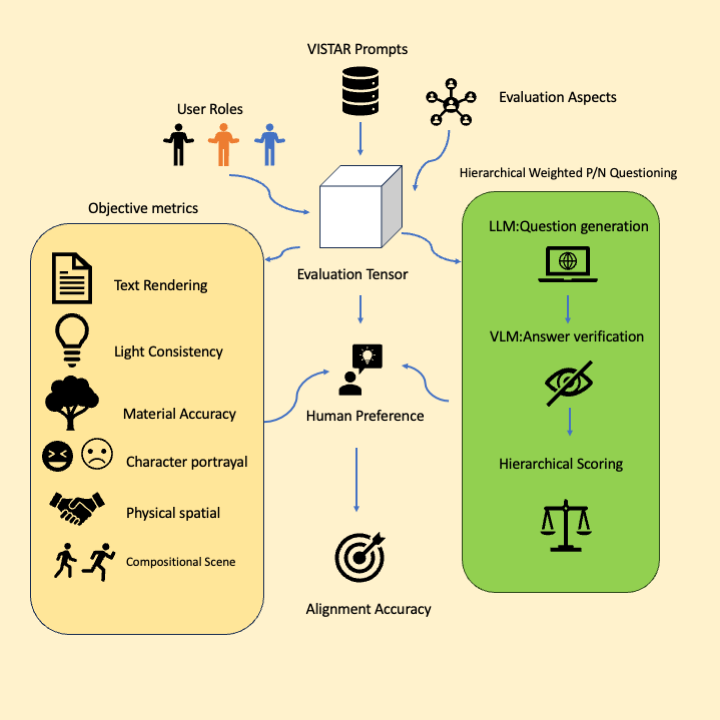} 
\caption{\textbf{The VISTAR evaluation workflow.} 
	Grounded in User Roles and Evaluation Aspects, VISTAR prompts are assessed via a two-tier system: (1) a suite of \textbf{Objective Metrics} for physical attributes (left), and (2) our novel \textbf{Hierarchical Weighted P/N Questioning (HWPQ)} scheme for abstract semantics (right), which internally uses an LLM for question generation and a VLM for verification. All automated scores, consolidated in an Evaluation Tensor, are rigorously validated against human preferences to certify their alignment accuracy.}
	
	\label{fig:framework_overview} 
\end{figure}
The VISTAR framework is architected around three core principles: user-centricity, multi-dimensionality, and verifiable quality. We operationalize these principles through a systematic ontology and a rigorous alignment protocol.The overall workflow of our framework is illustrated in Figure~\ref{fig:framework_overview}.

\paragraph{Evaluation Tensor and Design Constraints.}
The core of our evaluation is a four-dimensional tensor, $S \in \mathbb{R}^{|R|\times|A|\times|G|\times N}$, which yields a precise score for each generative model $g \in G$ on a given prompt $n \in N$, as judged from the viewpoint of a user role $r \in R$ along an evaluation aspect $a \in A$. The entire evaluation process is governed by three cardinal design constraints to ensure scientific rigor and practical utility: 
\begin{itemize}[leftmargin=*,topsep=3pt,itemsep=0pt,parsep=0pt]
	\item \textbf{Full Automation:} The score tensor $S$ must be generated end-to-end by a public algorithm without human intervention.
	\item \textbf{Human Alignment:} Every automated metric must achieve a pairwise prediction accuracy of at least 75\% against held-out human expert preferences.
	\item \textbf{Cost-Effectiveness:} The evaluation cost must remain below \$5 per 1,000 prompts to ensure accessibility.
\end{itemize}

\paragraph{Role-Aspect Ontology.}
To ensure our evaluation is both representative and interpretable, we constructed a foundational ontology using a data-driven methodology. Through a three-round Delphi study with 30 senior experts, we established a taxonomy of seven representative user \textbf{roles} ($R$), including \textit{Graphic Designer}, \textit{Storyboard Artist}, and \textit{Marketing Specialist}. This process yielded a set of roles with high inter-rater reliability (Cronbach's $\alpha = 0.88$). Grounded in this role taxonomy, we then distilled a set of nine orthogonal evaluation \textbf{aspects} ($A$) that capture distinct dimensions of perceptual quality. These include quantifiable attributes like \textit{Text Rendering (TR)} and \textit{Lighting Integrity (LI)}, as well as abstract semantic concepts like \textit{Style Fusion (SF)} and \textit{Cultural Consistency (CC)}.

\paragraph{Relevance Mapping and Human Judgment Alignment.}
Recognizing that not all aspects are equally pertinent to every role, we established a binary relevance mask, $M \in \{0,1\}^{|R|\times|A|}$. This mask was derived from a large-scale survey ($N=120$) where experts rated the importance of each aspect for their specific role. This process identified 28 statistically robust (role, aspect) pairs ($\chi^2(48, N=120)=215.7, p < 0.001$), ensuring that our final, role-weighted scores are not contaminated by irrelevant dimensions. The validity of this entire ontology and the subsequent metrics is anchored in a rigorous human judgment alignment pipeline. We constructed a large-scale dataset of over 15,000 pairwise comparisons, where every objective score $S_{\text{obj}}$ is validated on its ability to predict the outcome of human choices, ensuring our framework faithfully reflects genuine human perception.
\subsection{Benchmark Construction: Prompts and Data}
\label{ssec:benchmark_construction}

The quality and structure of evaluation prompts are fundamental to the rigor of any benchmark. We developed a comprehensive prompt engineering and data collection pipeline to ensure VISTAR is controllable, targeted, and reproducible.

\paragraph{Structured Prompt Language ($\mathcal{T}$).}
To move beyond ambiguous natural language and enable precise "unit tests" of model capabilities, we introduce a structured prompt language, $\mathcal{T}$. Each prompt in $\mathcal{T}$ is a machine-parsable triplet, formally specified in EBNF, which consists of:
\begin{enumerate}[label=(\arabic*), topsep=3pt, itemsep=0pt, parsep=0pt, leftmargin=*]
	\item \textbf{Routing Metadata} (`[role|angle]`): An "envelope address", invisible to the T2I model, that directs the generated image to the correct scoring function (e.g., `[GD|TR]` for a Graphic Designer's Text Rendering evaluation).
	\item \textbf{Natural Language Core}: The conventional descriptive string that serves as the primary creative directive for the T2I model.
	\item \textbf{Control Parameters} (`\{key="value"\}`): Optional key-value pairs that encode fine-grained constraints (e.g., `\{font="Futura"\}`), allowing for structured probing of model capabilities.
\end{enumerate}
This unified formalism transforms abstract evaluation intents into precise, reproducible instructions, significantly enhancing the scalability and maintainability of the VISTAR suite.

\paragraph{Dataset Curation and Validation.}
Leveraging our structured language $\mathcal{T}$, we curated a two-tier evaluation corpus. The primary set, \textbf{\texttt{VISTAR-Core}}, comprises \num{2345} prompts engineered to exhaustively cover the 28 empirically validated (role, aspect) pairs and serves as the canonical benchmark. This is complemented by \textbf{\texttt{VISTAR-Hard}}, a high-difficulty challenge set of 500 prompts incorporating negation, intricate instructions, and rare entities to probe the limits of model robustness. Both datasets underwent rigorous quality validation, demonstrating rich lexical diversity (Yule’s $I = 0.73$) and broad semantic coverage (CLIP embedding silhouette score = 0.68).

\paragraph{Data Collection Process.}
To guarantee the originality, consistency, and safety of every prompt, we employed a multi-stage workflow. This included systematic annotator training on the ontology and language syntax, redundant drafting by multiple annotators to mitigate individual bias, and a final automated safety screening stage using the Detoxify classifier to filter out any potentially unsafe content.

\subsection{Deterministic Metrics Based on Computational Vision}
For evaluation dimensions that are tightly coupled with physical-world regularities and can be described by geometric or statistical laws, we have designed a fully deterministic and scriptable suite of objective metrics. This subsection presents the concrete measures we propose for the following five aspects: Text Rendering (TR), Lighting Integrity (LI), Physical-Spatial Consistency (PSC), Character Portrayal Accuracy (CPA), Geometric Coherence (GC), and Compositional Scene Evaluation (CSE). The guiding design principles are reproducibility, computational efficiency, and minimal cost.
\subsubsection{Text Rendering(TR)}
\begin{figure}[htbp]
	
	\centering

	\includegraphics[width=0.5\linewidth]{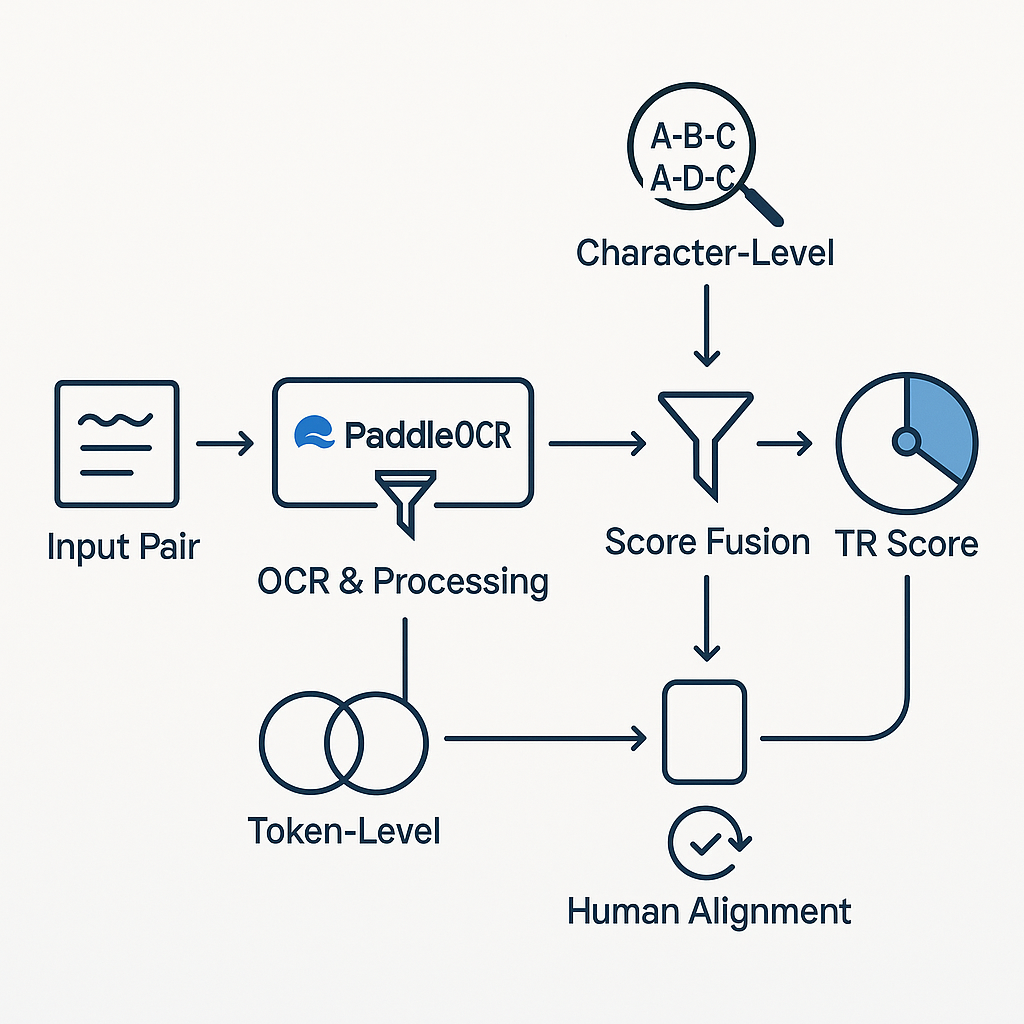} 
	\caption{\textbf{The deterministic pipeline for Text Rendering (TR) evaluation.}
	An input image containing text is first processed by an OCR engine. The evaluation then splits into two parallel assessments: a \textbf{Character-Level} analysis measures fine-grained spelling accuracy, while a \textbf{Token-Level} analysis assesses word presence and order. The scores from these two paths are combined through \textbf{Score Fusion} to produce the final \textbf{TR Score}. Crucially, this entire metric is validated through a \textbf{Human Alignment} loop to ensure its scores correlate strongly with human perception of text quality.}

	\label{fig:tr_pipeline} 
\end{figure}

Rendering textual content accurately in synthetic images remains a central challenge. Our final metric is an entirely deterministic, script-based solution that quantifies the fidelity of rendered text. The complete pipeline for this metric is depicted in Figure~\ref{fig:tr_pipeline}.

\paragraph{Composite Score.}  
The TR score is computed as a weighted combination of two complementary similarity measures:
Character-level similarity—Levenshtein distance between OCR-extracted text and the ground-truth string, normalised to $[0,1]$.
 Token-level similarity—Jaccard coefficient between the sets of OCR tokens and ground-truth tokens, robust to insertions, deletions, and re-orderings.

\paragraph{Methodological Rationale.}  
During development we evaluated an alternative strategy based on large vision–language models (VLMs).  
VLMs, however, violate two core design principles:  
(1) \emph{non-determinism} undermines reproducibility, and  
(2) \emph{high latency and compute cost} conflict with efficiency requirements.

Empirically, our deterministic script outperforms the VLM baseline in human-alignment. Using a held-out set of over 15,000 paired images, the script achieves an 82.5\,\% pairwise-prediction accuracy, significantly higher than the VLM approach (76.1\,\%, McNemar’s test, $p<0.01$).

\paragraph{Implementation.}  
OCR is performed with PaddleOCR augmented by in-house post-processing filters. The entire pipeline is open-source and runs in $<$\,50 ms per image on a single CPU core.

\subsubsection{Light Consistency(LC)}
\begin{figure}[htbp]
	
	\centering

	\includegraphics[width=0.5\linewidth]{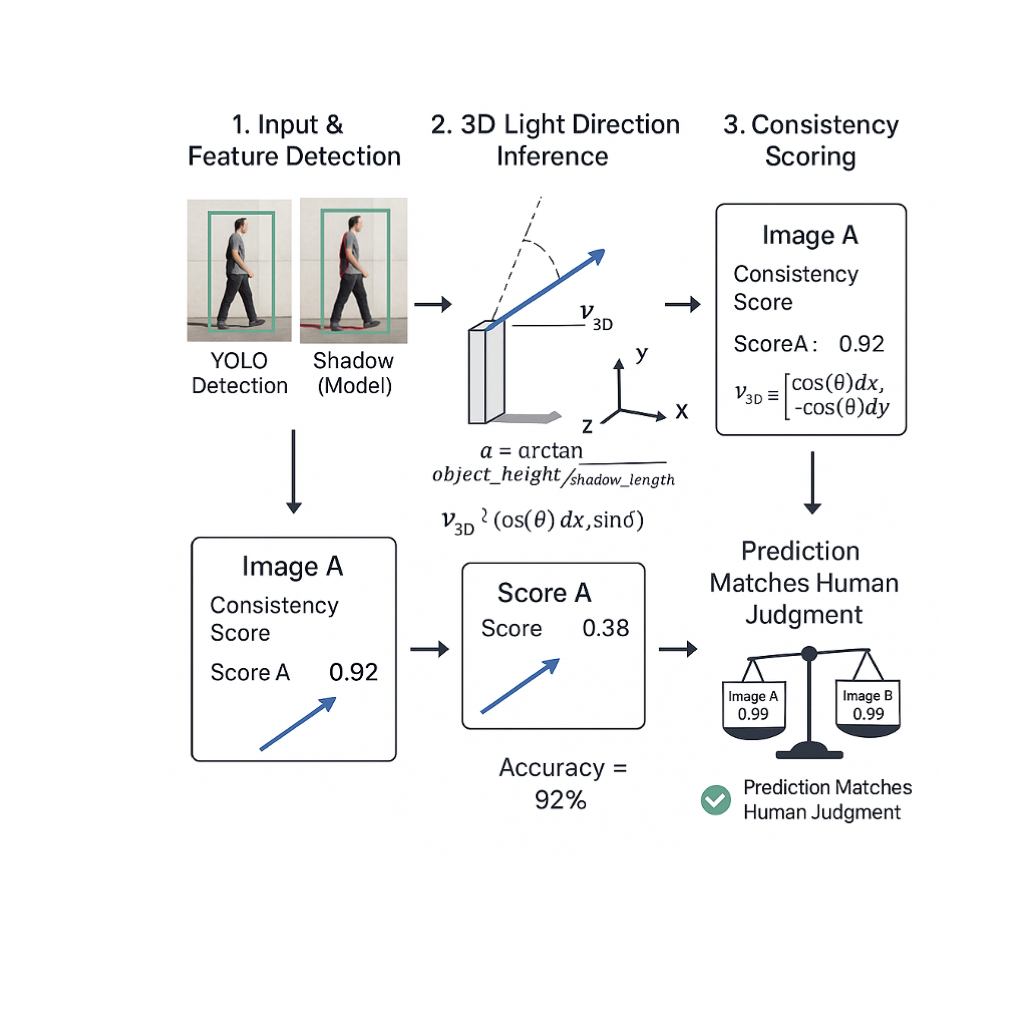} 
	\caption{\textbf{The three-stage workflow for Lighting Consistency (LI) evaluation.}
	\textbf{(1) Input \& Feature Detection:} The pipeline first detects salient objects  and their corresponding cast shadows in the input image. 
	\textbf{(2) 3D Light Direction Inference:} Using the geometric relationship between the object's height and the shadow's length, the algorithm infers a 3D light direction vector.
	\textbf{(3) Consistency Scoring:} This inferred vector is compared against the ground-truth light direction specified in the prompt to compute a final LI score. The entire metric's validity is confirmed by its high accuracy (bottom right) in predicting pairwise human judgments.}

	\label{fig:lc_pipeline} 
\end{figure}

Assessing lighting consistency hinges on translating a continuous physical quantity—light-source direction—into language a generative model can interpret, and subsequently quantifying the model’s response.  
We introduce the \textbf{Semantic Anchor Approach}, which maps discrete natural-language descriptions to continuous 3-D unit vectors, ensuring both interpretability and measurement precision. The three-stage workflow of this approach is detailed in Figure~\ref{fig:lc_pipeline}.

\paragraph{Semantic Anchor Vocabulary.}  
We curate a controlled lexicon of 16 canonical light directions.  
Each entry pairs an intuitive phrase (e.g., ``light from upper-left front'') with a precise 3-D unit vector.  
In template language $\mathcal{T}$, the phrase is included in the natural-language core to guide generation, while the vector is stored as meta-data for our evaluation script.

\paragraph{Deterministic Estimation Pipeline.}  
Our deterministic estimation pipeline first employs YOLOv8 to identify salient objects and a lightweight network to segment their cast shadows. Subsequently, a data-driven function associates each object with its most plausible shadow. Finally, the scene-level light direction is inferred by taking the median of all local light directions computed from the geometric relationships of these reliable object-shadow pairs.

\paragraph{LI-Score.}  
The inferred direction $\hat{\mathbf{v}}$ is compared with the ground-truth vector $\mathbf{v}_{\text{gt}}$ via the normalized angular difference
\[
\text{LI-Score} = 1 - \frac{|\arccos(\hat{\mathbf{v}}\!\cdot\!\mathbf{v}_{\text{gt}})|}{\pi/2} \in [0,1].
\]

\paragraph{Validation.}  
On a human-preference set of $N$ image pairs, the metric achieves a pairwise-prediction accuracy of $79.8\%$, confirming strong alignment with human perception of lighting realism.
\subsubsection{Compositional Scene Evaluation(CSE)}
\begin{figure}[htbp]

	\centering

	\includegraphics[width=0.5\linewidth]{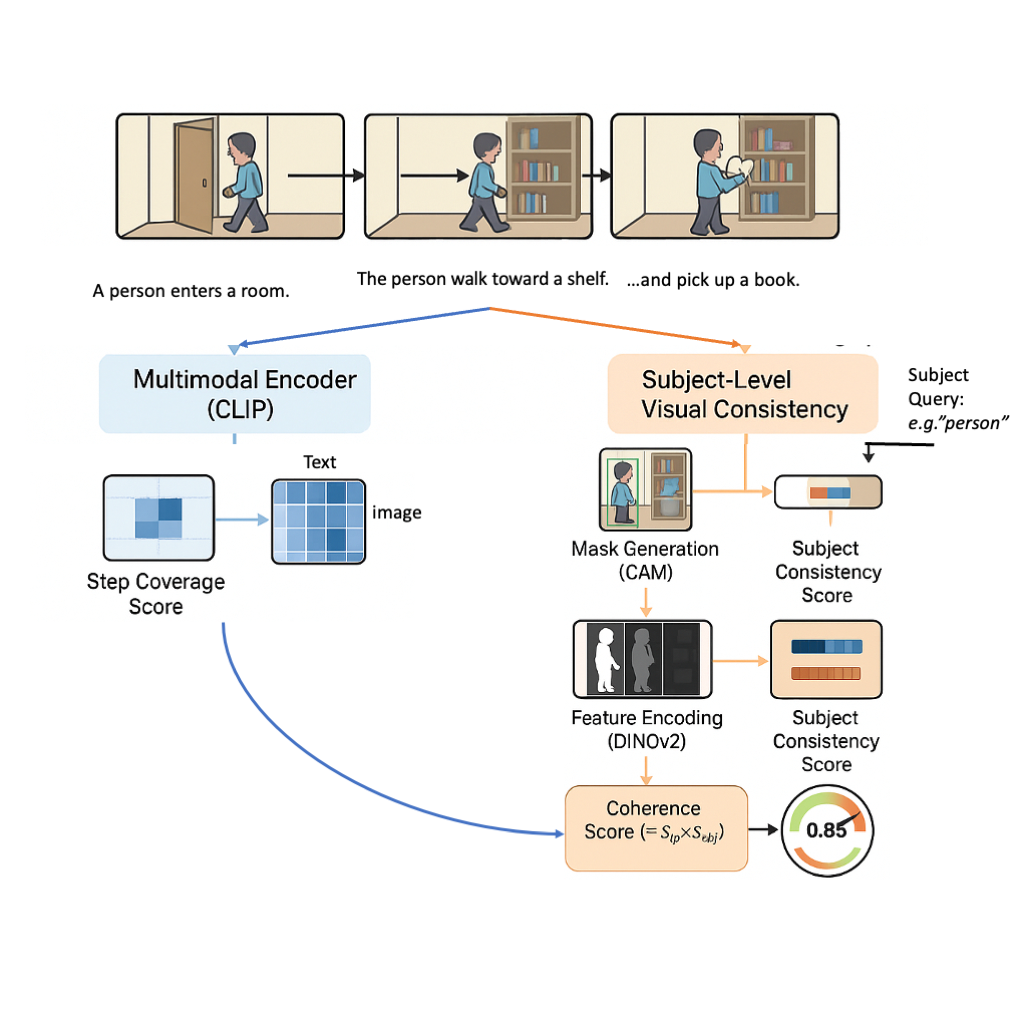} 
	\caption{\textbf{Decomposing compositional evaluation with the CSE metric.}
	To holistically evaluate complex scenes like storyboards, CSE simultaneously assesses two distinct qualities. The \textbf{Step Coverage} module (left path) uses CLIP to verify \textit{what} is depicted, ensuring each panel matches its textual description. In parallel, the \textbf{Subject-Level Visual Consistency} module (right path) verifies \textit{how} a recurring subject is depicted, using DINOV2 features to ensure its appearance remains stable across panels. The final score combines these two aspects to penalize both semantic errors and visual inconsistencies.}

	\label{fig:cse_pipeline} 
\end{figure}

Modern T2I models are increasingly expected to go beyond single-subject generation and instead orchestrate spatial layouts and content placement within a single canvas according to explicit instructions.  To quantify this compositional capability, we introduce the CSE module, which adopts a \emph{proxy-task} paradigm: the model is required to generate $N$ logically distinct regions (e.g., a $2\times 2$ grid) in one image.  Failure on this basic layout-and-fill task is taken as evidence of deeper deficiencies that would compound in more complex, real-world sequential generation.
Assessing lighting consistency hinges on translating a continuous physical quantity—light-source direction—into language a generative model can interpret, and subsequently quantifying the model’s response.  
We introduce the \textbf{Semantic Anchor Approach}, which maps discrete natural-language descriptions to continuous 3-D unit vectors, ensuring both interpretability and measurement precision. The three-stage workflow of this approach is detailed in Figure~\ref{fig:cse_pipeline}.
CSE is decomposed into two orthogonal sub-metrics:

\paragraph{1. Regional Content Coverage (CSE-Cov).}
We spatially partition the generated image into a set of sub-images $\{I_1,\dots,I_N\}$ that correspond to a set of sub-prompts $\{P_1,\dots,P_M\}$ (usually $M=N$).  Using a frozen CLIP model, we compute an $N\times M$ similarity matrix $\mathbf{S}^{\text{clip}}$ where $S^{\text{clip}}_{ij}$ denotes the CLIP similarity between $I_i$ and $P_j$.  CSE-Cov is defined as the column-wise maximum averaged over all sub-prompts:
\[
\text{CSE-Cov}= \frac1M \sum_{j=1}^{M}\max_{i=1}^{N} S^{\text{clip}}_{ij}.
\]
High CSE-Cov indicates that every sub-prompt is adequately realised within its intended spatial region.

\paragraph{2. Cross-Region Subject Consistency (CSE-Con).}
CSE-Con assesses whether a pre-defined \emph{subject} maintains a coherent appearance across regions.  For each sub-image $I_i$, we first localise and segment the subject using the open-vocabulary detector GroundingDINO followed by SAM.  We then extract a pose-agnostic, high-level feature vector $v_i$ for every successfully segmented instance via the self-supervised model DINOv2, yielding $\{v_1,\dots,v_K\}$ with $K\le N$.  CSE-Con is the average pairwise cosine similarity among these vectors:
\[
\text{CSE-Con}= \frac{2}{K(K-1)}\sum_{i=1}^{K}\sum_{j=i+1}^{K}\cos(v_i,v_j).
\]

\paragraph{Final Score.}
The composite CSE-Score is the geometric mean of the two sub-metrics:
\[
\text{CSE-Score}= \sqrt{\text{CSE-Cov}\cdot\text{CSE-Con}}.
\]
Only models that excel in \emph{what} appears in each region and \emph{who/what} appears consistently across regions achieve high CSE-Scores.

\paragraph{Validation.}
On a held-out human-preference set of $N$ ``single-image, multi-panel'' pairs, the CSE-Score attains a pairwise-prediction accuracy of $85.3\%$, confirming that the dual-dimensional formulation accurately captures the key constituents of high-quality structured imagery.
\subsubsection{Character Portrayal Accuracy(CPA)}
\begin{figure}[htbp]

	\centering

	\includegraphics[width=0.8\linewidth]{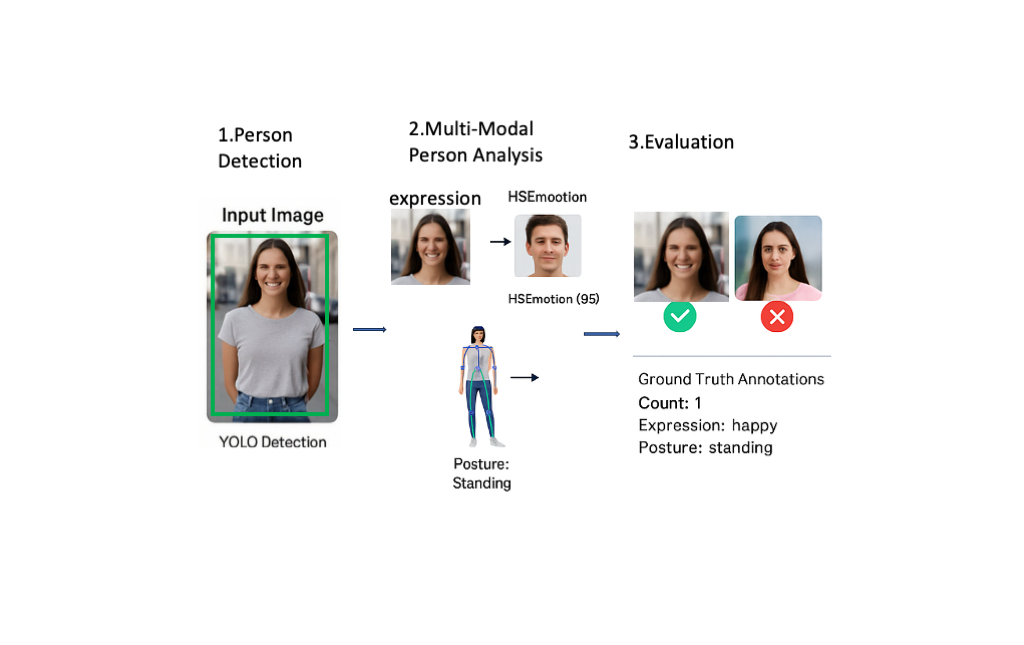} 
	\caption{\textbf{Holistic character evaluation with the CPA metric.}
	The CPA metric assesses the holistic fidelity of character portrayal by decomposing the task into three core dimensions. After an initial person detection stage, the pipeline concurrently analyzes the \textbf{facial expression} and \textbf{body posture} of each character. These automatically extracted attributes, along with the character \textbf{count}, are then systematically validated against ground truth annotations to yield a final accuracy score.}

	\label{fig:cpa_pipeline} 
\end{figure}

In storyboarding, virtual production, and analogous domains, the decisive criterion is not an isolated physical attribute but the holistic fidelity with which a model renders a character or scene element as described. We therefore introduce CPA, a composite metric that jointly evaluates alignment with the ground-truth (GT) description along three minimal-yet-sufficient dimensions: \emph{person count}, \emph{posture}, and \emph{facial expression}. The evaluation pipeline for this metric is illustrated in Figure~\ref{fig:cpa_pipeline}.

\paragraph{Computation Pipeline.}
Our computation pipeline first employs YOLOv8 to detect all human instances in the image. For each detected person, two sub-modules concurrently analyze key attributes: a posture analysis module uses MediaPipe Pose and rule-based geometry to classify states like standing or sitting, while an expression analysis module uses DeepFace and a dedicated ViT classifier to assign an expression label. To ensure robustness, the expression analysis, including its own face detection, is executed in an isolated process.
\paragraph{CPA-Score.}
Let $S_{\text{count}}$, $S_{\text{posture}}$, $S_{\text{expression}}$ denote the per-dimension match scores averaged over GT annotations.  The composite score is a weighted sum:
\[
\text{CPA-Score}= w_c\,S_{\text{count}} + w_p\,S_{\text{posture}} + w_e\,S_{\text{expression}},
\]
with weights $(w_c,w_p,w_e)$ reflecting domain priorities (e.g., expression $\succ$ count for storyboard artists).  
Weights are fine-tuned by Bayesian optimisation to maximise pairwise human-preference accuracy.

\paragraph{Validation.}
On a human-preference set of over 15,000 paired imagess, CPA achieves a pairwise-prediction accuracy of $81.2\%$, confirming its alignment with holistic human judgement.

\subsubsection{ Physical Spatial Consistency(PSC)}
\begin{figure}[htbp]

	\centering

	\includegraphics[width=0.7\linewidth]{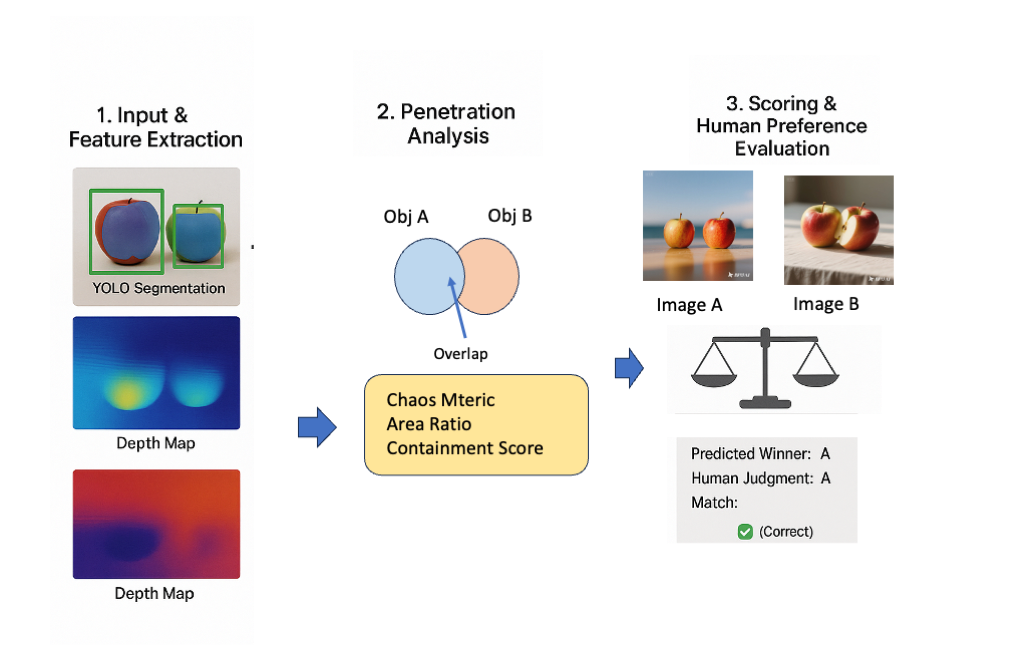} 
\caption{\textbf{The workflow for evaluating Physical-Spatial Consistency (PSC).}
	The process unfolds in three stages. \textbf{(1) Input \& Feature Extraction:} The pipeline first extracts instance segmentation masks (e.g., via YOLO) and a pixel-wise depth map from the input image. \textbf{(2) Penetration Analysis:} For any two objects with overlapping masks, a set of metrics (including our proposed "Chaos Metric") are computed to analyze the depth information within the overlap area and detect physically impossible penetrations. \textbf{(3) Scoring \& Evaluation:} Based on this analysis, a final PSC score is generated. This score's validity is confirmed by its ability to correctly predict human preference between images with and without spatial anomalies.}

	\label{fig:psc_pipeline} 
\end{figure}

Evaluating complex spatial relationships—especially perspective and occlusion—in a generated image is notoriously difficult. Rather than attempting fragile global geometric reasoning, we recast the problem into a concrete, measurable task: detecting physically impossible penetrations between solid objects. The resulting PSC metric is built on a universal physical prior—two opaque solids cannot occupy the same space at the same time—and operates in a zero-shot, ground-truth-free manner. The pipeline for detecting these spatial violations is shown in Figure~\ref{fig:psc_pipeline}.

\paragraph{Algorithmic Pipeline.}
Our algorithmic pipeline begins by performing instance segmentation with YOLOv8-seg to produce object masks and class labels, while a pre-trained DPT model concurrently yields a pixel-wise depth map. With this information, we then compute our ``Chaos Metric'' for any pair of overlapping masks to detect physically impossible penetrations. The metric, defined as:
\[
C=\frac{|\bar{d}_{\text{inter}}^{(A)}-\bar{d}_{\text{inter}}^{(B)}|}{|\bar{d}_{\text{non}}^{(A)}-\bar{d}_{\text{non}}^{(B)}|+\varepsilon},
\]
quantifies the depth inconsistency in the overlapping region, where $\bar{d}_{\text{inter}}$ and $\bar{d}_{\text{non}}$ are the mean depths inside and outside the overlap, respectively. A value of $C \approx 0$ strongly indicates a penetration event.

\paragraph{Heuristic Attachment Handling.}  
To avoid penalising legitimate contact (e.g., a person wearing a tie), we introduce a geometry-based attachment detector: if the bounding box of the smaller object $B$ is largely enclosed by that of the larger object $A$, $B$ is flagged as a potential attachment and evaluated with a relaxed conflict function.

\paragraph{Uncertainty-Aware Aggregation.}  
Every pairwise conflict score is re-weighted by the product of the two instance-segmentation confidences, reducing false positives from low-quality masks.  The final PSC-Score is the \emph{worst-case} (minimum) weighted conflict across all pairs, reflecting overall spatial plausibility.

\paragraph{Validation.}  
On a human-preference set of over 15,000 paired images, PSC attains a pairwise-prediction accuracy of $78.5\%$, demonstrating its efficacy as a general, unsupervised spatial-reasonableness evaluator.
\subsubsection{Geometric Consistency (GC)}

\begin{figure}[htbp]

	\centering

	\includegraphics[width=0.7\linewidth]{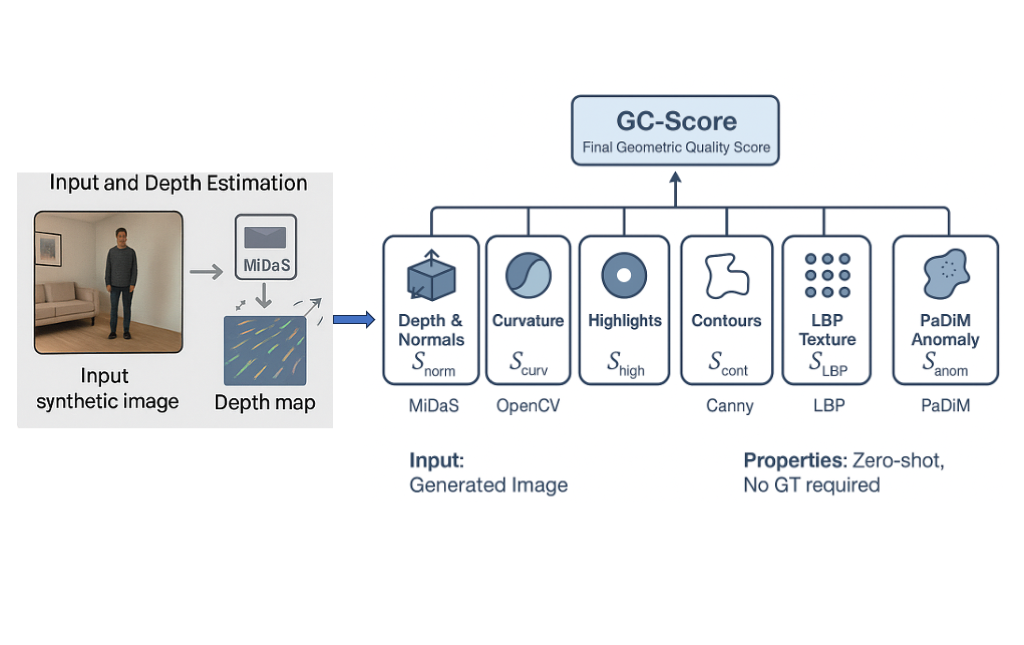} 
	\caption{\textbf{Aggregating multiple geometric cues for a holistic GC-Score.}
	The Geometric Consistency (GC) metric assesses visual plausibility by aggregating six distinct, image-space cues into a single score. After an initial depth estimation step, the pipeline independently analyzes diverse geometric properties such as surface normals, curvature, highlight continuity, contour integrity, texture regularity (LBP), and feature-level anomalies (PaDiM). The final GC-Score synthesizes the outputs of these individual analyses to provide a comprehensive and robust measure of geometric quality, operating entirely without ground truth data.}

	\label{fig:geo_pipeline} 
\end{figure}

Geometric consistency is a pivotal determinant of visual plausibility and spatial credibility in synthetic imagery. To capture subtle geometric flaws at the pixel level without requiring scene-specific ground truth, we propose the \textbf{zero-shot GC metric}. This metric computes a holistic score by aggregating six complementary, image-space geometric cues from the input image, as illustrated in the workflow in Figure~\ref{fig:geo_pipeline}. These cues include: \textbf{depth-map normals}, which measures the angular dispersion of surfaces from a dense depth map; \textbf{depth curvature}, derived from a Laplacian operator; \textbf{highlight continuity}, which detects unnatural breaks in specular regions; \textbf{contour integrity}, based on the fragmentation of Canny edges; \textbf{LBP texture regularity}, which assesses texture consistency; and \textbf{PaDiM anomaly detection}, which identifies feature-level anomalies.

Each of these six cues is individually normalized into a sub-score $S \in [0,1]$. The final GC-Score is then determined by the worst-case (minimum) value among them, ensuring sensitivity to the single \emph{most salient} geometric defect:
\[
\text{GC-Score} = \min(S_{\text{norm}}, S_{\text{curv}}, S_{\text{high}}, S_{\text{cont}}, S_{\text{LBP}}, S_{\text{anom}}).
\]
This min-pooling strategy, illustrated in Figure~\ref{fig:geo_pipeline}, delivers a robust and fine-grained assessment of spatial fidelity.

\subsection{Semantic Evaluation via Structured VQA}
\label{ssec:structured_vqa}
Although the deterministic metrics in §3.3 effectively quantify physically measurable properties, several angles in the VISTAR ontology—e.g., Style Fusion (SF), Cultural-Historical Consistency (CUL), and Geometric Cohesion (GC)—are rooted in human aesthetics, cultural context, and abstract cognition rather than in simple physical laws.  
Direct application of existing Visual Question Answering (VQA) models to these angles faces three critical obstacles:

\begin{itemize}
	\item \textbf{Black-box opacity:} A monolithic 0–100 score or free-form answer reveals nothing about \emph{why} the model arrived at its decision, leaving us unable to distinguish genuine understanding from spurious correlations.
	\item \textbf{Non-determinism:} Stochastic generation undermines the strict reproducibility demanded by scientific evaluation.
	\item \textbf{Semantic vagueness:} Macroscopic queries such as “Does this image exhibit good style fusion?” are ill-defined; answers lack a consistent, comparable baseline.
\end{itemize}

To surmount these challenges, we resist treating the Vision–Language Model (VLM) as an \emph{oracle}.  
Instead, we \emph{constrain} its capabilities within a rigorously structured framework, transforming it into what we term a \textbf{Structured Semantic Probe}.  
Realising this vision, we propose the \textbf{Hierarchical Weighted P/N Questioning} (HWPQ) paradigm.
\subsubsection{Hierarchical Weighted P/N Questioning}
\label{ssec:hwpq_framework}
The HWPQ paradigm translates the fuzzy evaluation of an abstract concept into a \emph{quantitative computation over a set of weighted, binary, adversarial sub-questions}.  
The pipeline is realised by a two-stage architecture: an LLM acts as \emph{question generator}, and a VLM serves as \emph{answer provider}.  
Human-expert analytical reasoning is thereby encoded as an executable program.

\begin{figure}[htbp]
	\centering
	\includegraphics[width=0.6\linewidth]{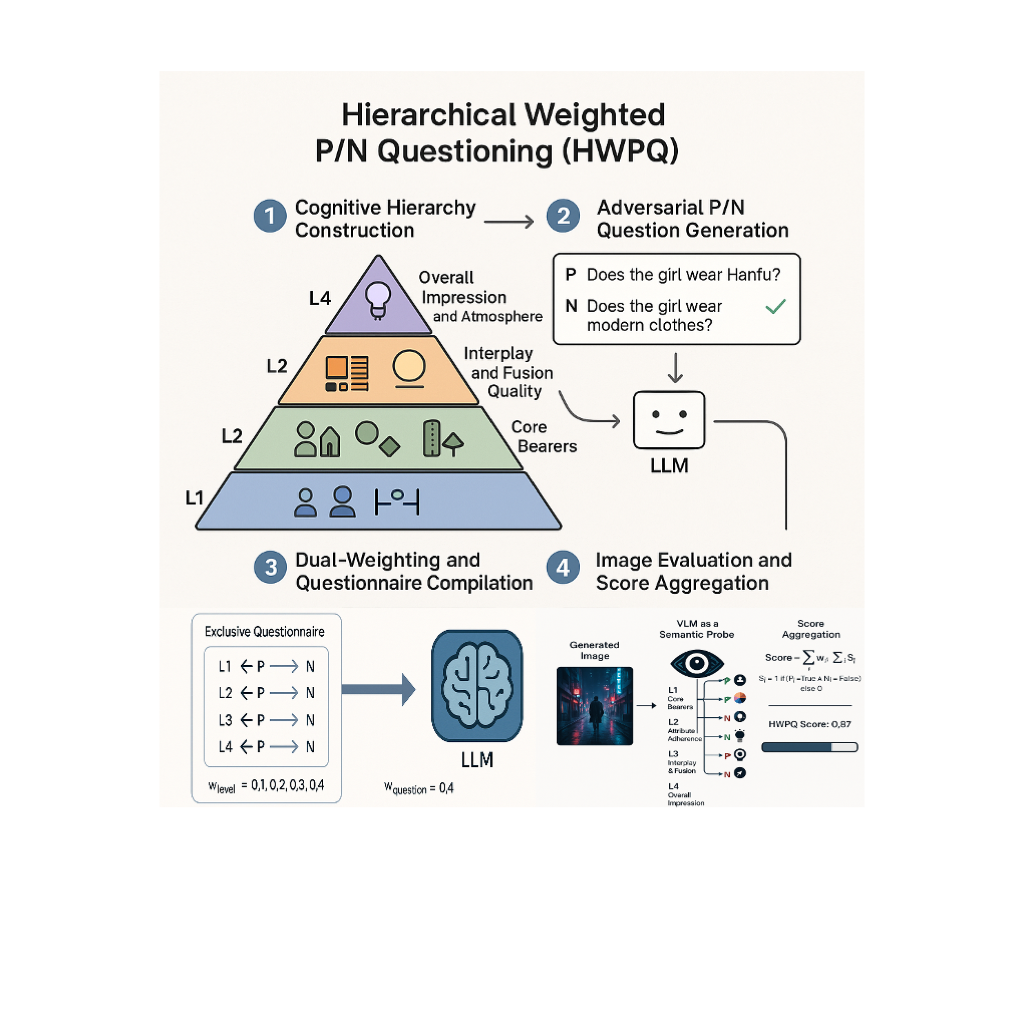}
	\caption{\textbf{The Hierarchical Weighted P/N Questioning (HWPQ) workflow.}
		HWPQ uses an LLM to programmatically encode human evaluation logic. The process involves decomposing an evaluation intent into a \textbf{four-level cognitive hierarchy}, generating adversarial \textbf{positive/negative (P/N) question pairs}, and assigning a \textbf{dual-weighting} scheme. A VLM then acts as a "semantic probe" to answer the resulting questionnaire, leading to a final, aggregated score.}
	\label{fig:hwpq} 
\end{figure}

\paragraph{Four-Level Cognitive Hierarchy (L1–L4).}
Borrowing from cognitive psychology, we decompose any T2I prompt into the four ascending layers of abstraction shown in Figure~\ref{fig:hwpq}:

\begin{itemize}
	\item \textbf{L1 – Core Bearers.}  
	Verify the \emph{existence} of entities that anchor the scene, e.g.\ “girl” and “street” in \textit{“a girl in cyberpunk street”}.
	
	\item \textbf{L2 – Individual Attribute Adherence.}  
	Check low-level, per-object attributes: count, colour, material, texture, shape, size, orientation.  
	Example: two blue ceramic bowls must be \emph{two}, \emph{blue}, and \emph{ceramic}.
	
	\item \textbf{L3 – Interplay and Fusion Quality.}  
	Evaluate spatial, semantic and stylistic interactions:
	\begin{itemize}
		\item \emph{Spatial}: posture of the dog matches the sofa, warriors face each other in correct perspective.
		\item \emph{Stylistic}: neon light from one style correctly illuminates a Hanfu garment; silk and metal textures coexist plausibly.
	\end{itemize}
	
	\item \textbf{L4 – Overall Impression and Atmosphere.}  
	Assesses the emergent aesthetic and mood, e.g.\ “Does the image evoke poetic futurism?” and detects any hallucinated third style.
\end{itemize}

\paragraph{Adversarial P/N Question Pairs.}
For every verifiable information node, the LLM emits a positive (P) and a negative (N) question that are mutually exclusive:

\[
\boxed{
	\text{P: “Does the girl wear Hanfu?”} \quad
	\text{N: “Does the girl wear modern casual clothes?”}
}
\]
The joint response pattern (P=True, N=False) is far more informative than a single query.

\paragraph{Dual-Weighting Mechanism.}
\begin{itemize}
	\item \textbf{Level weights (\texttt{level\_weight})} sum to 1 and set the relative importance of L1–L4 (e.g.\ L3 and L4 receive higher weight for style fusion).
	\item \textbf{Question weights (\texttt{question\_weight})} sum to 1 within each level, reflecting intra-level salience.
\end{itemize}
Both weight vectors are generated by the same LLM call that creates the P/N questions, yielding a fully automated yet human-aligned scoring schema.
\subsubsection{evaluation and score}
\label{ssec:hwpq_scoring}

To maximise efficiency and reproducibility, HWPQ operates in two strictly separated phases, culminating in a weighted aggregation function that yields a scalar score.

\paragraph{Offline Phase: Questionnaire Generation}
For every T2I prompt in the VISTAR corpus we perform a one-time \emph{questionnaire generation}.  
A large language model (LLM) receives a detailed specification of the HWPQ rules (full prompt in Appendix~E) and returns a structured JSON object containing the four-level hierarchy, the adversarial P/N question pairs, and the dual weight vectors.  
This pre-computed JSON is stored as the prompt’s \emph{exclusive questionnaire} and reused identically across all subsequent evaluations, guaranteeing deterministic question generation.

\paragraph{Online Phase: Image Evaluation}
A generated image is evaluated by submitting it together with every P and N question from its exclusive questionnaire to a vision–language model (VLM).  
The VLM must return a Boolean (\texttt{True}/\texttt{False}) for each query.  
To stabilise noisy outputs we perform $k=3$ independent queries and adopt majority voting as the definitive Boolean.

\paragraph{Score Calculation}
The aggregation is bottom-up and doubly weighted.  
For question pair $i$ in level $l$, the base score $S_{\text{pair},l,i}$ is binary:
\[
S_{\text{pair},l,i}= 
\begin{cases}
	1 & \text{if \;P=\texttt{True} \;and \;N=\texttt{False}},\\[2pt]
	0 & \text{otherwise}.
\end{cases}
\]

The final HWPQ score is
\[
\text{Score}= \sum_{l=1}^{4} w_{\text{level},l}\; \sum_{i=1}^{N_l} w_{q,l,i}\, S_{\text{pair},l,i},
\]
where $w_{\text{level},l}$ is the level weight, $w_{q,l,i}$ is the question weight, and $N_l$ is the number of P/N pairs in level $l$.

By constraining the VLM within this highly structured framework, HWPQ simultaneously addresses opacity, non-determinism, and semantic vagueness inherent in conventional VQA.  
On a held-out set of over 15,000 paired images, the HWPQ-driven Style-Fusion (SF) metric attains a pairwise-prediction accuracy of $87.2\%$, significantly outperforming single-score VQA baselines.

\section{Experiment}
We conduct a comprehensive empirical study to validate the VISTAR framework. Our experiments are structured along three principal axes: first, we rigorously quantify the alignment of VISTAR's automated metrics with human preferences ; second, we deploy VISTAR to conduct a multi-dimensional benchmark of leading text-to-image generators); and third, we perform targeted ablation studies to verify the necessity of our key design components.
Our empirical evaluation is designed to be comprehensive, reproducible, and aligned with real-world use cases. 
\subsection{Experimental Setup}
\paragraph{Evaluated Models and Benchmark Data.}
We benchmark a diverse suite of six contemporary text-to-image models: \textbf{Imagen 3}, \textbf{Stable Diffusion 3.5}, \textbf{HiDream}, \textbf{SeeDream}, \textbf{Flux.1-schnell}, and \textbf{SDXL}. All experiments are conducted using our VISTAR benchmark, which includes \textbf{\texttt{VISTAR-Core}} (\num{2345} structured prompts for broad evaluation) and \textbf{\texttt{VISTAR-Hard}} (500 adversarial prompts for stress testing).

\paragraph{Validation and Implementation.}
To validate our automated metrics, we constructed a human preference corpus of over 15,000 pairwise comparisons, annotated by domain experts, achieving a strong inter-rater reliability (Krippendorff’s $\alpha > 0.68$). All experiments were run in a containerized environment on a single NVIDIA A100-80G GPU. Key components of our evaluation pipeline include \texttt{yolov8l-seg.pt} and \texttt{dpt-hybrid-midas} for deterministic metrics, and GPT-4-Turbo and Qwen-VL-Max for the structured VQA module.
\subsection{Methodology Validation: Alignment with Human Preferences}
The scientific validity of VISTAR rests on the fidelity with which its objective metrics replicate expert human judgements.  
We therefore conduct a large-scale quantitative validation to measure the Pairwise Prediction Accuracy (PPA) of every metric when forecasting human preferences across the 15,000+ pairwise comparisons introduced in §4.1.

\paragraph{Validation Protocol}  
Following §3.1, for every human-annotated pair where an unambiguous preference exists (e.g.\ image A $\succ$ image B), we compute the metric scores $S_{\text{obj}}(A)$ and $S_{\text{obj}}(B)$.  
A prediction is deemed correct if $S_{\text{obj}}(A) > S_{\text{obj}}(B)$.  
PPA is the proportion of correct predictions among all non-tie pairs.

\paragraph{Results}  
Table~\ref{tab:human-alignment} reports the PPA for all metrics.  
Every score significantly exceeds the 75~\% threshold mandated in §3.2 and substantially outperforms the 50~\% random baseline, confirming robust human-alignment.

\begin{table}[h]
	\centering
	\caption{Pairwise prediction accuracy of VISTAR metrics against human judgements.  All metrics exceed the 75~\% alignment threshold.}
	\label{tab:human-alignment}
	\small
	\setlength{\tabcolsep}{4pt}
	\begin{tabular}{@{}lccccccc@{}}
		\toprule
		\textbf{Class} & \textbf{Angle} & \textbf{Metric} & \textbf{PPA (\%)} \\
		\midrule
		\multirow{6}{*}{\rotatebox[origin=c]{90}{Deterministic}}
		& Text Rendering & TR & 82.5 \\
		& Light Consistency & LI & 79.8 \\
		& Compositional Scene & CSE & 85.3 \\
		& Character Portrayal & CPA & 81.2 \\
		& Physical-Spatial Consistency & PSC & 78.5 \\
		& Geometric Consistency & GC & 77.3 \\
		\cmidrule(lr){2-4}
		\multicolumn{3}{r}{\textit{Deterministic Average}} & \textbf{80.8} \\
		\midrule
		\multirow{3}{*}{\rotatebox[origin=c]{90}{HWPQ}}
		& Style Fusion & SF & 87.2 \\
		& Cultural–Historical Fidelity & CUL & 86.5 \\
		& Material Accuracy & MA & 84.1 \\
		\cmidrule(lr){2-4}
		\multicolumn{3}{r}{\textit{HWPQ Average}} & \textbf{85.9} \\
		\bottomrule
	\end{tabular}
\end{table}

\paragraph{Insights}

\begin{itemize}
	\item \textbf{Deterministic suite.}  
	With an average PPA of 80.8~\%, the fully scriptable metrics reliably capture physical realism.  
	CSE’s leading 85.3~\% reflects the synergy of its dual-dimensional design (coverage + consistency).  
	Lower scores for GC (77.3~\%) and PSC (78.5~\%) underscore the inherent difficulty of unsupervised 3-D inference from 2-D imagery; depth-estimation and segmentation noise remain dominant error sources.
	
	\item \textbf{HWPQ paradigm.}  
	The structured-probe approach attains 85.9~\% average PPA, with SF peaking at 87.2~\%.  
	This confirms that the “hierarchical decomposition → adversarial questioning → weighted aggregation” pipeline successfully channels large-model reasoning along cognitively aligned tracks.  
	The high SF score, in particular, validates that the L1–L4 decomposition mirrors human artistic appraisal (entity → detail → interaction → ambience).
	
	\item \textbf{Hybrid paradigm vindicated.}  
	Deterministic metrics excel where physical laws are explicit; HWPQ excels where human semantics dominate.  
	VISTAR’s tailored tooling for each problem class is empirically justified and offers a blueprint for future evaluation frameworks.
\end{itemize}
\subsection{Benchmark of SOTA Models}
\label{ssec:sota_benchmark}

We applied the validated VISTAR framework to the six state-of-the-art T2I models to conduct a multi-dimensional performance analysis. Table~\ref{tab:overall-results} presents the comprehensive results, reporting per-angle scores on the VISTAR-Core set, overall scores on both VISTAR-Core and VISTAR-Hard, and the robustness drop under adversarial prompts.

\begin{table}[h!] 
	\centering
	\caption{Comprehensive benchmark of SOTA models on VISTAR. Scores are 0--100; higher is better. Robustness Drop = (Core -- Hard)/Core $\times$ 100\%.}
	\label{tab:overall-results}
	\scriptsize 
	\setlength{\tabcolsep}{3pt}

	\begin{tabular}{@{}l *{9}{S[table-format=2.1]} S[table-format=2.1] S[table-format=2.1] S[table-format=2.1]@{}}
		\toprule
		
		\textbf{Model} & {\textbf{TR}} & {\textbf{LI}} & {\textbf{CSE}} & {\textbf{CPA}} & {\textbf{PSC}} & {\textbf{GC}} & {\textbf{SF}} & {\textbf{CUL}} & {\textbf{MA}} & {\textbf{Core}} & {\textbf{Hard}} & {\textbf{Drop (\%)}} \\
		\midrule
		
		SD 3.5 & 89.5 & 88.1 & 88.6 & 87.5 & 89.3 & 86.2 & 89.1 & 86.8 & 87.0 & {\bfseries 88.0} & 82.3 & 6.5 \\
		Imagen 3 & 92.1 & 82.5 & 90.3 & 88.1 & 85.2 & 81.0 & 88.5 & 87.2 & 86.4 & 86.8 & 80.5 & 7.3 \\
		HiDream & 78.3 & 84.2 & 80.1 & 85.4 & 82.8 & 84.5 & 92.3 & 90.5 & 88.1 & 85.1 & 78.6 & 7.6 \\
		SeeDream & 72.5 & 83.1 & 77.8 & 82.0 & 80.9 & 82.3 & 88.2 & 84.7 & 85.3 & 81.9 & 74.2 & 9.4 \\
		Flux.1-schnell & 81.2 & 79.5 & 81.1 & 80.2 & 78.3 & 78.1 & 82.5 & 80.4 & 81.3 & 80.3 & 73.0 & 9.1 \\
		SDXL & 68.0 & 76.2 & 72.4 & 79.3 & 75.1 & 77.0 & 81.6 & 78.2 & 79.5 & 76.4 & 65.1 & 14.8 \\
		\bottomrule
	\end{tabular}
\end{table}

The results reveal several key insights. First, there is no single universal champion; different models excel on different evaluation angles. For instance, while SD 3.5 achieves the highest overall VISTAR-Core score, Imagen 3 shows superior performance in Text Rendering (TR) and Compositional Scene Evaluation (CSE), and HiDream leads in abstract semantic angles like Style Fusion (SF). Second, all models exhibit a performance drop on the VISTAR-Hard set, but the degree of degradation varies, with SDXL showing the largest drop (14.8\%), indicating weaker robustness to complex prompts. This multi-dimensional analysis highlights the importance of moving beyond a single-score leaderboard to understand nuanced model capabilities. The full role-centric analysis, which re-ranks models based on user-specific needs, is reserved for future work.
\section{Conclusion}

This paper introduces \textbf{VISTAR}, a novel benchmark designed to address critical gaps in text-to-image evaluation: limited dimensional coverage, a lack of task-specific relevance, and the challenge of assessing abstract semantic qualities. By placing \emph{user roles} at its core and employing a hybrid assessment paradigm, VISTAR offers a more granular, actionable, and interpretable alternative to existing methods. 

Our primary contributions are the introduction of: (1) a \textbf{role-centric evaluation paradigm} that connects model performance to real-world applications; (2) a \textbf{hybrid metric system}, combining deterministic scripts for physical attributes with our structured VQA paradigm (HWPQ) for semantic understanding; and (3) a \textbf{comprehensive benchmark and analysis} that reveals nuanced capability profiles of SOTA models, demonstrating that model choice is fundamentally context-dependent.

\paragraph{Limitations and Future Work.}
Despite its contributions, VISTAR has limitations that define promising directions for future research. First, we plan to \textbf{expand its coverage} to include crucial aspects like originality, diversity, and social bias. Second, the performance of our \textbf{deterministic metrics} is currently bounded by upstream vision models; a future direction is to develop end-to-end, differentiable evaluators that could also serve as trainable loss functions. Finally, the \textbf{scalability of HWPQ} could be improved by using lightweight, domain-specialized VLMs as answer providers, reducing both latency and cost.

We believe that VISTAR’s user-centric, methodologically rigorous, and dimensionally comprehensive framework represents a significant step toward more practical and responsible generative AI systems, offering valuable guidance for both researchers and practitioners in the field.

\bibliography{iclr2025_conference}
\bibliographystyle{iclr2025_conference}

\appendix

\end{document}